\documentclass[letterpaper, 10 pt, conference]{ieeeconf}  

\usepackage{siunitx}
\usepackage{graphicx} 

\usepackage{lipsum}
\usepackage{ upgreek }

\usepackage{booktabs}
\usepackage{multirow}
\usepackage{tabularx}
\usepackage{rotating}
\usepackage{bigstrut}

\usepackage{color, colortbl}
\IEEEoverridecommandlockouts                              

\usepackage[maxfloats=256]{morefloats}
\makeatletter
\let\NAT@parse\undefined
\makeatother
\usepackage[bookmarks=false, linkcolor=blue, urlcolor=blue, citecolor=blue]{hyperref} 
\usepackage{fancyhdr}
\hypersetup{
    colorlinks=true,
    linkcolor=red,
    filecolor=magenta,      
    urlcolor=blue,
    pdfstartview={FitH},
    citecolor =blue
    }

\title{\LARGE \bf
Hybrid Gripper with Passive Pneumatic Soft Joints for\\Grasping Deformable Thin Objects 
}

\author{Ngoc-Duy Tran$^{\dagger1}$, Hoang-Hiep Ly$^{\dagger1}$, Xuan-Thuan Nguyen$^{1}$, Thi-Thoa Mac$^{1}$, Anh Nguyen$^{2}$, and Tung D. Ta$^{3}$
\thanks{$^{\dagger}$Co-first authors contributed equally.}
\thanks{*This work was supported by JSPS Grant-in-Aid for Scientific Research~(B) Grant Number 23H01376, Japan and the Royal Society ISPF International Collaboration Awards 2023 (Japan).}
\thanks{$^{1}$School of Mechanical Engineering, Hanoi University of Science and Technology, Vietnam
        {\tt\small duy.tn206101@sis.hust.edu.vn, \{hiep.lyhoang, nguyenxuan, thoa.macthi\}@hust.edu.vn}}%
\thanks{$^{2}$Department of Computer Science, University of Liverpool, UK
        {\tt\small anh.nguyen@liverpool.ac.uk}}%
\thanks{$^{3}$Graduate School of Information Science and Technology, The University of Tokyo, Japan 
        {\tt\small tung@csg.ci.i.u-tokyo.ac.jp}}%
}

\begin{document}

\maketitle
\thispagestyle{empty}
\pagestyle{empty}

\begin{abstract}

Grasping a variety of objects remains a key challenge in the development of versatile robotic systems. The human hand is remarkably dexterous, capable of grasping and manipulating objects with diverse shapes, mechanical properties, and textures. Inspired by how humans use two fingers to pick up thin and large objects such as fabric or sheets of paper, we aim to develop a gripper optimized for grasping such deformable objects. Observing how the soft and flexible fingertip joints of the hand approach and grasp thin materials, a hybrid gripper design that incorporates both soft and rigid components was proposed. The gripper utilizes a soft pneumatic ring wrapped around a rigid revolute joint to create a flexible two-fingered gripper. Experiments were conducted to characterize and evaluate the gripper performance in handling sheets of paper and other objects. Compared to rigid grippers, the proposed design improves grasping efficiency and reduces the gripping distance by up to eightfold.

\end{abstract}


\section{INTRODUCTION}  \label{sec: introduction} 
Gripper is a crucial component in most current robotic applications~\cite{shintake2018soft, Hughes2016-jg}. The grippers enable robots to interact with their surroundings by contacting and grasping objects~\cite{Hughes2016-jg}. The design of robotic grippers that can perform complex tasks with objects of various shapes, sizes, and materials with high performance while ensuring the necessary safety for humans and objects is a challenge in robotic research~\cite{Carbone2013-pl, Achilli2020-hn, Billard2019-gl}.

Most conventional grippers used in industrial robotic systems are rigid grippers. These grippers are constructed from rigid joints and links. Actuators, such as motors, can be installed directly on the gripper joints or links to generate gripper movements~\cite{Liu2020-wd, Nishimura2018-vz}. Furthermore, some grippers used cables or tendons activated by motors to grasp and manipulate objects~\cite{Do2023-ds, Ko2020-cw}. The tendon-driven or cable-driven gripper tends to be more compact and flexible than motor-activated grippers~\cite{Wei2023-yi}. Rigid grippers are highly sturdy and stable, making them suitable for tasks that require high precision, repeated actions, or the handling of heavy objects~\cite{Long2020-yq, Samadikhoshkho2019-hf}. However, a drawback of rigid grippers is their difficulty in applications that require handling irregularly shaped objects. In addition, when handling soft or fragile objects, rigid grippers have a high possibility of damaging them.

Soft grippers show greater flexibility in grasping various shaped objects, compared to conventional rigid grippers~\cite{Ham2018-vf}. In addition, soft grippers have been shown to be safe for handling fragile items without causing damage and are safe for humans because they are made of soft materials~\cite{Chen2021-wi}. The movement of soft grippers can be activated by various mechanisms, such as shape memory alloys~\cite{Kim2016-zy}, cable-driven systems~\cite{Khatib2016-ra}, granular jamming~\cite{Gotz2022-cg}, and pneumatic mechanisms being the most common~\cite{Hao2016-ug}. Despite showing many advantages, soft grippers also have several drawbacks, such as weak grip force (which leads to difficulty in holding heavy objects or keeping them for a long time), low bandwidth control, and difficulty in modeling or controlling precisely in real-time as rigid grippers~\cite{AboZaid2024-db, AliAbbasi2024-yb}.

\begin{figure}
    \centering
    \includegraphics[width=\linewidth]{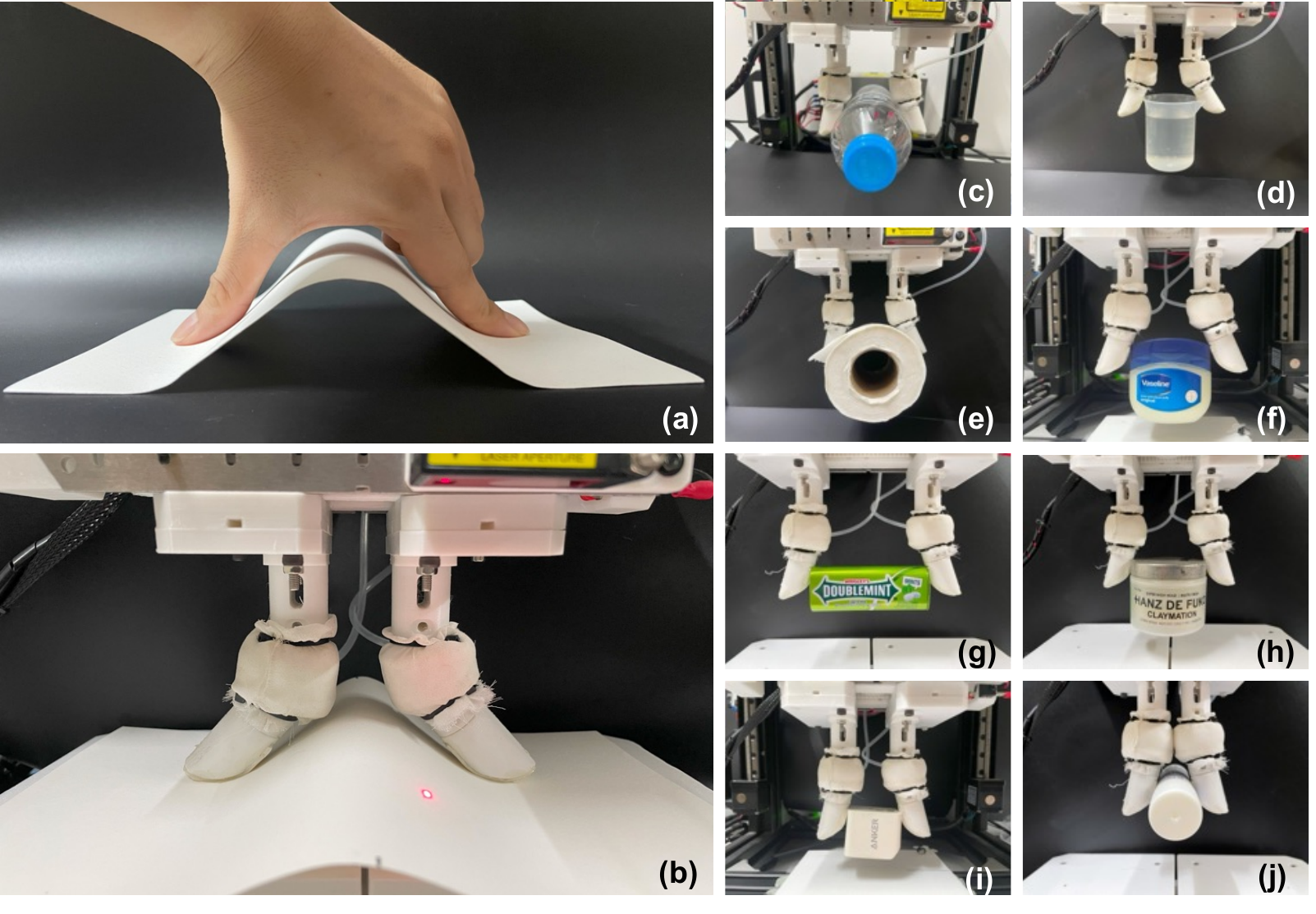}
    \caption{Hybrid gripper with soft pneumatic rings is used to grasp deformable objects, thin objects, and a variety of other objects. (a) The way humans use finger hand to pick up a sheet of paper; the rest is our gripper grasping, (b) a sheet of paper, (c) a PET bottle, (d) a water-filled plastic cup, (e) a paper roll, (f) a plastic box, (g) a gum box, (h) a cylinder Box, (i) a mobile charger, (j) a glue stick.}
    \label{fig: Introduction}
\end{figure}

\begin{figure} 
    \centering
    \includegraphics[width=\linewidth]{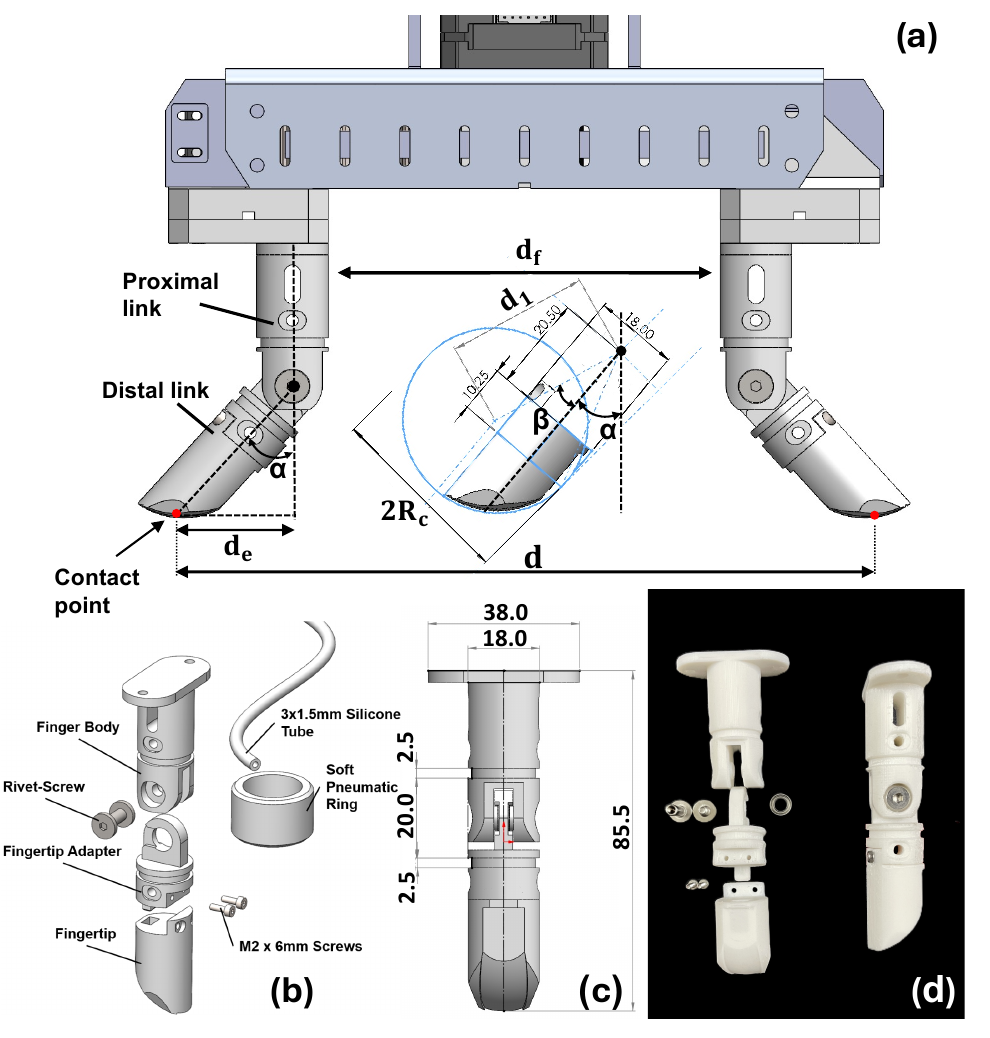}
    \caption{ Hybrid gripper design and prototype (a) Gripper including dimensions of two fingers in grasping operation and predetermined dimensions of the fingertip ($R_c=\SI{20}{\milli\meter}$, $d_1=\SI{31}{\milli\meter}$, $\beta=\SI{20.64}{\degree}$) with $d$ standing for the distance between two contact points, $d_f$ standing for the distance between two fingers and $d_e$ standing for the extended distance of bending fingertip. (b) Exploded view of the finger including 7 components: finger body, fingertip adapter, fingertip, rivet-screw, M2 screws connecting all rigid parts, soft pneumatic ring consisting of soft core and silicone tube. (c) The drawing view shows the dimension of the finger. (d) The final finger after fabricating and assembling.}
    \label{fig: finger and fingertip}
\end{figure}

Currently, there is a trend toward using hybrid grippers in robotic applications~\cite{Yi2018-hp, Zhou2020-pm}. Hybrid grippers combine soft components or actuators integrated into rigid links, allowing them to take advantage of being both rigid and soft grippers mentioned earlier. For example, these grippers have high passive adaptability to different shapes and sizes~\cite{Wang2017-iv} while still offering high efficiency for tasks that require strength and precision~\cite{Amend2012-gs}. However, existing hybrid grippers are still limited in terms of dexterity, the ability to handle lateral forces, and the gripper size~\cite{Zhou2017-lu}. To address these issues, some studies have proposed hybrid grippers with pneumatic-based soft components attached between the rigid links of the gripper~\cite{Amend2012-gs, Zhou2017-lu}. With this design, the movement and stiffness of the gripper joints can be flexibly controlled. However, grasping some objects that are larger than the gripper’s opening angle, such as a sheet of paper, remains challenging.

This study was inspired by the human fingers grasping a paper sheet larger than the hand palm, as shown in Fig.~\ref{fig: Introduction}(a). For a flat, thin object of this size, the hand spreads and presses on the surface of the paper, causing the distal phalanges of the index finger and thumb to bend laterally. After pressing with moderate force, the angle between the index finger and thumb gradually closes, lifting the paper. This study hypothesized that bendings of the distal phalanx and stiffness of the distal interphalangeal (DIP) joints are key factors that allow the human hand or gripper to pick up large or thin objects. To verify this hypothesis, a hybrid gripper prototype is developed. The gripper links are made of rigid components. A soft component is attached to the gripper's finger joint to vary the stiffness of the DIP joints, and the stiffness was controlled by air pressure. The effects of DIP joint stiffness and distal link bending angle on grasping various objects, including large objects like papers, were evaluated in this study.


\section{Hybrid Gripper Design and Prototype} \label{sec: method}

\subsection{Gripper fingers} \label{subsec: design and fabrication hard link}
In this study, a gripper with two gripper fingers was designed and developed. The design was inspired by the human finger, with one gripper finger representing the thumb (left finger) and the other gripper finger (right gripper finger) representing the index finger, as shown in Fig.~\ref{fig: finger and fingertip}(a).

The gripper in this study is simplified from human finger anatomy so that each finger consists of two links: the proximal link (referred to as the finger body) and the distal link (which includes the fingertip and fingertip adapter), as shown in Figures~\ref{fig: finger and fingertip}(a) and~\ref{fig: finger and fingertip}(b). The fingertip is designed to resemble the distal phalanx of the human finger. The dimensions of the gripper fingertip are described in an inset in Fig.~\ref{fig: finger and fingertip}(a). The dimensions of the gripper finger are described in Fig.~\ref{fig: finger and fingertip}(c). The contact surface with the object has a curved shape to ensure a constant contact area when the gripper touches a flat surface. The proximal link and distal link can rotate relative to each other through a revolute joint. In this design, the fingertip can rotate \SI{80}{\degree} in both clockwise and counterclockwise directions around the revolute joint. Additionally, a thin silicone pad is attached to the fingertip to increase friction when grasping objects. The rigid links of the gripper were 3D printed with Polyethylene Terephthalate Glycol (PETG) material using a fused deposition modeling 3D printer (Neptune~4~Pro).

\subsection{Soft pneumatic ring} \label{subsec: soft pneumatic ring}

\begin{figure}
        \centering
    \includegraphics[width=\linewidth]{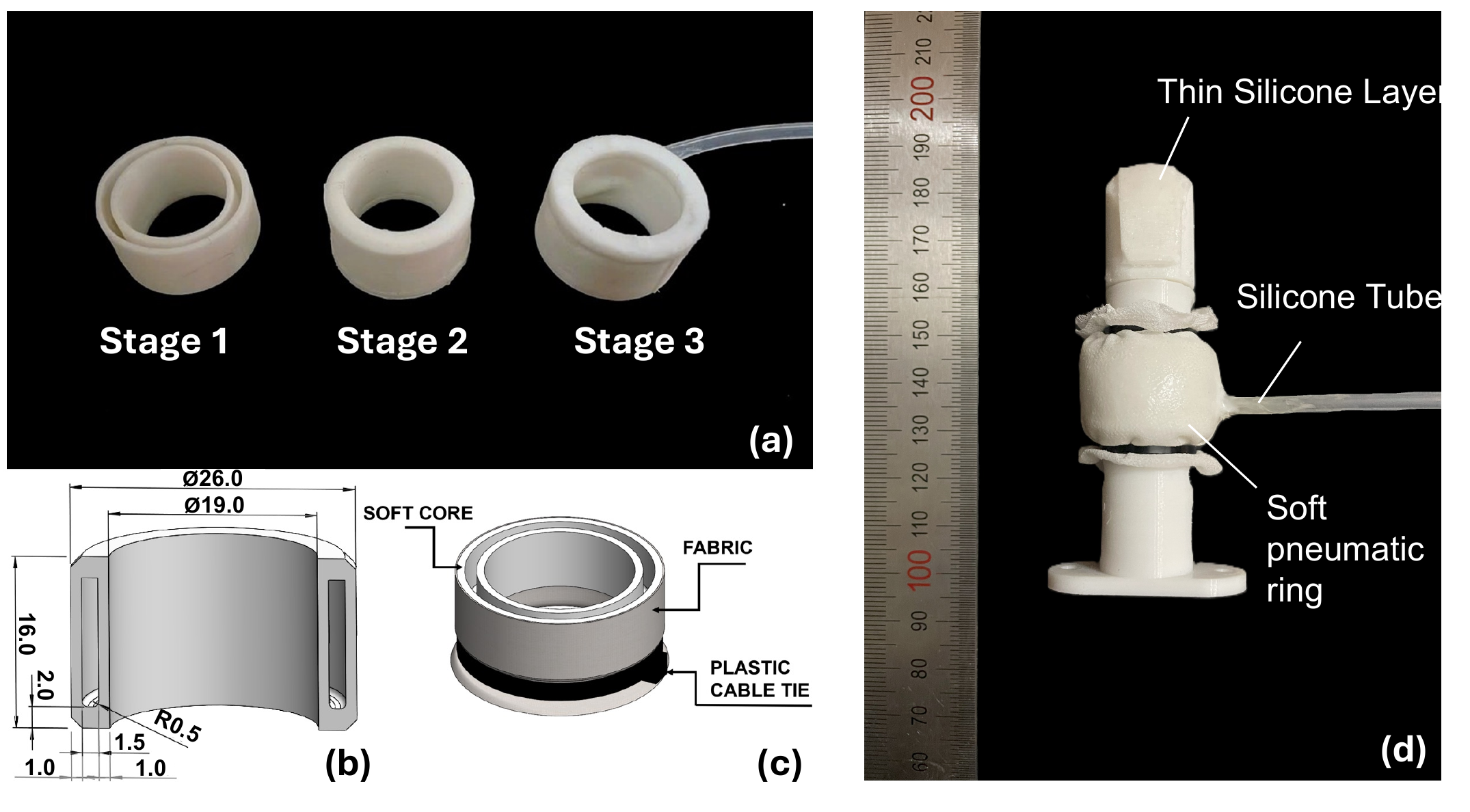}
    \caption{ Soft pneumatic ring. (a) Three stages of fabricating soft core including Stage 1 Main body Casting, Stage 2 Cap Closing, and Stage 3 Silicone tube inserting. (b) The drawing view shows the dimension of the cross-section soft core. (c) The components of the final soft pneumatic ring consist of soft core, a layer of fabric, and plastic cables. (d) The final finger consists of hard links and a soft pneumatic ring.}
    \label{fig: soft actuator and final finger}
\end{figure}

A soft component, called a soft pneumatic ring, is attached to the gripper to adjust the stiffness of the revolute joint at the gripper finger. The soft pneumatic ring has a length, inner diameter, and outer diameter of \SI{16}{\milli\meter}, \SI{19}{\milli\meter}, and \SI{26}{\milli\meter}, respectively, as shown in Fig.~\ref{fig: soft actuator and final finger}(b). Inside the soft pneumatic ring, there is a concentric cavity with a width of \SI{1.5}{\milli\meter} and a height of \SI{12}{\milli\meter}. A fabric layer, \SI{0.5}{\milli\meter} thick, is glued to the outside of the soft pneumatic ring, as shown in Fig.~\ref{fig: soft actuator and final finger}(c). Pneumatic pressure is applied to the cavity of the soft pneumatic ring through a silicone tube with an inner diameter of \SI{1.5}{\milli\meter} and an outer diameter of \SI{3}{\milli\meter}. The fabric layer ensures that the soft pneumatic ring will hardly stretch outward but only inward. As the pressure increases, the soft pneumatic ring compresses the gripper joint, increasing the joint's stiffness.

The soft pneumatic ring is fabricated using a molding process. First, male and female molds are designed and fabricated using a 3D printer. Then, white silicone rubber, with a viscosity of \SI{25000}{\milli\pascal.\second}, is poured into the mold. The soft pneumatic ring with an open cavity in stage~1 (Fig.~\ref{fig: soft actuator and final finger}(a)) is created, and the hardness after curing is Shore~25A. In stage~2, another mold is created to seal the open part of the cavity from stage 1 with the same white silicone rubber. In stage~3, a small piece of fabric is glued to the outside of the soft pneumatic ring using the same white silicone rubber as in the previous stages. Finally, a small hole is made in the body of the soft pneumatic ring, allowing the silicone tube to be attached, which is used to input pneumatic pressure into the soft pneumatic ring cavity.

\subsection{Gripper prototype} \label{subsec: Gripper Prototype}

After the soft pneumatic ring is fabricated, it is installed into the gripper finger, as shown in Fig.~\ref{fig: soft actuator and final finger}(d). Two soft pneumatic rings are fitted into the two gripper fingers and securely attached to the gripper's frame. The movement of the two gripper fingers is controlled by a Nema17 stepper motor (Usongshine~17HS4401) mounted on top of the frame. When the motor rotates, both gripper fingers move inward (flexion) or outward (extension) simultaneously, with the same speed and displacement. The range of distance - $d_{f}$ (as shown in Fig.~\ref{fig: finger and fingertip}(a)) between the two gripper fingers varies in the range of \qtyrange{0}{100}{\milli\meter}. The speed and rotation angle of the motor is controlled by a stepper motor driver circuit (MKS~Gen~L~V1.0).

The stiffness of the soft pneumatic ring is regulated by the pneumatic pressure supplied to it. Air pressure from an air compressor passes through an electro-pneumatic regulator (SMC~ITV1030-212BL50-Xl88) before reaching the soft pneumatic rings. The air pressure delivered to the soft pneumatic rings can be adjusted by varying the input voltage (\qtyrange{0}{5}{\volt}).

\begin{figure}[t]
    \centering
    \includegraphics[width=1\linewidth]{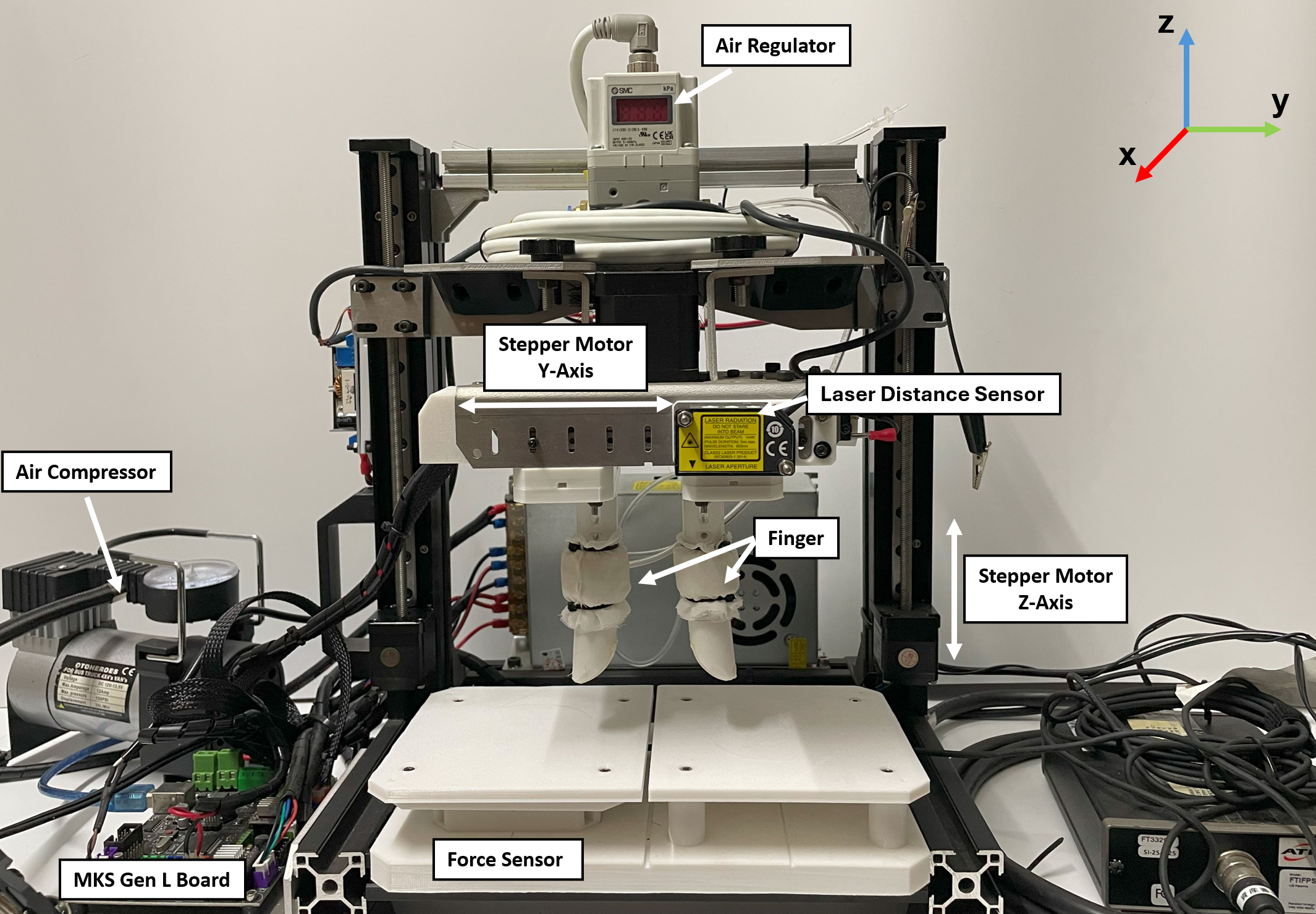}
    \caption{Grasping experimental system using the hybrid gripper.}
    \label{fig: full system}
\end{figure}

\section{Experiments}  \label{sec: Experiments}
In this study, an experimental system was set up to measure the fundamental characteristics of the gripper and evaluate its performance in grasping various objects, as shown in Fig.~\ref{fig: full system}. The proposed gripper was mounted on a steel frame, which was attached to two linear stages capable of synchronized movement. The up-and-down movement along the z-axis was controlled by two \SI{24}{\volt} stepper motors. All the components were installed on aluminum frames. The movement range of the gripper was measured by a microlaser distance sensor (Panasonic~HG-C1200 - \SI{200}{\micro\meter} repeat accuracy) mounted on the gripper frame.
\subsection{Soft joint stiffness experiments } \label{subsec: Fundamental experiments}
\begin{figure}
    \centering
    \includegraphics[width=1\linewidth]{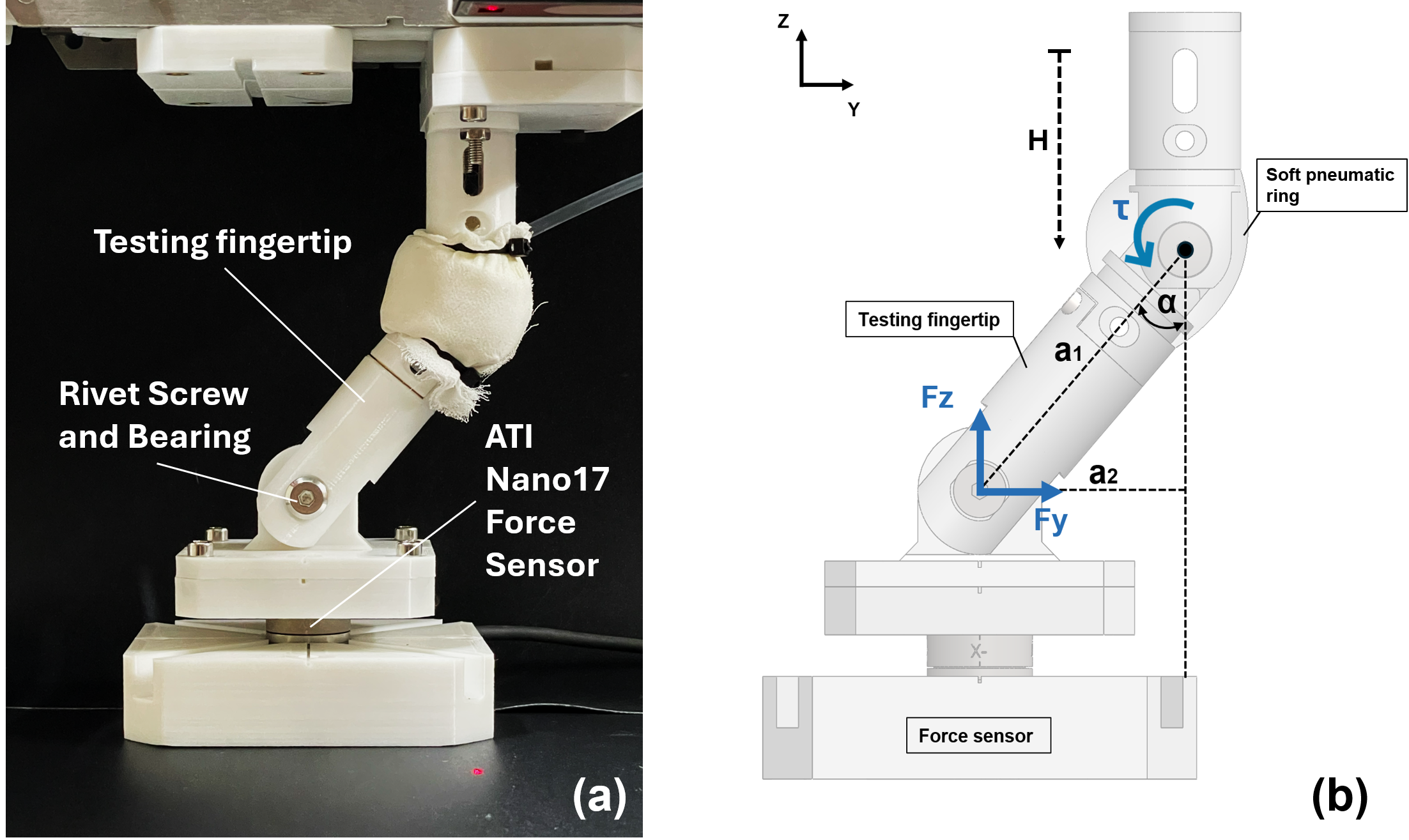}
    \caption{Soft joint stiffness experiment. (a) A picture of the actual system, (b) experiment devices and force applied on testing fingertip. $F_z \text{ and } F_y$ are the force exerted from the joint of the force sensor device, and $\uptau$ is the torque generated from the soft ring to the testing fingertip. Based on this, the torque is calculated by Eq.~\ref{eq: torque experiment 1}.}
    \label{fig: Experiment 1}
\end{figure}
This experiment was set up to evaluate the effect of the rotation angle and stiffness of the gripper joint when increasing the air pressure supplied to the soft ring. The resisting torque at the gripper joint generated during the distal link bending backward was used for evaluation.

To minimize the effect of slipping, which may occur when pressing the fingertip onto the surface of the experimental system's base, a testing fingertip with a length of $a_{1}=\SI{50}{\milli\meter}$ was substituted into the gripper finger, as shown in Fig.~\ref{fig: Experiment 1}(a) and~\ref{fig: Experiment 1}(b). The testing fingertip is connected to the base through a revolute joint. A commercial 6-axis Force/Torque sensor (ATI~Nano~17) is fixed underneath the base to measure the interaction forces of the gripper on the base. In this experiment, the gripper will only apply forces within the $yz$ plane, so the two forces $F_{y}$ and $F_{z}$ are used for calculation and evaluation, as shown in Fig.~\ref{fig: Experiment 1}(b). The bending angle~$\alpha$ between the gripper fingertip and the gripper finger body will be adjusted by controlling the displacement~$H$ along the z-axis of the gripper body. The relationship between $H$ and $\alpha$ is expressed as:
\begin{equation}
 H = a_1(1-\cos{(\alpha)})
\end{equation}
At each $\alpha$ angle position, the air pressure supplied to the soft pneumatic ring will vary. As the air pressure increases, the resisting torque on the revolute joint of the gripper also changes. This resisting torque ($\uptau$) can be estimated using the following equation,
\begin{equation}
    \uptau = F_za_1\sin(\alpha)-F_y a_1 \cos(\alpha)
    \label{eq: torque experiment 1}
\end{equation}
In this experiment, the $\alpha$ angle will be varied in the range of~\qtyrange{0}{80}{\degree} with the stepping of~\SI{5}{\degree}. The air pressure supplied to the soft pneumatic ring will be in the range of~\qtyrange{0}{150}{\kilo\pascal} with stepping of \SI{10}{\kilo\pascal}. For each $\alpha$ angle and pressure value, the measurement was conducted 10 times, and the average value will be used for evaluation. The result of this experiment is discussed in section~\ref{sec:result-torque}.



\subsection{Grasping experiments} \label{subsec: Evaluation experiments}
In this experiment, the base, fabricated using a 3D printer, is securely mounted onto the aluminum frame, as shown in Fig.~\ref{fig: full system}. On top of the base are two separate flat plates. The right plate is fixed to the base, while the left plate is attached to the Force/Torque sensor (ATI~Nano~17) to measure the interaction forces of the gripper on the target objects.

\subsubsection{Thin-large object grasping experiment} \label{Thin object grasping experiments}
The hypothesis behind this new grasping method is that no sheet of paper can be completely flat; the paper will have a certain initial deflection on its surface. This means that the force applied to the paper through its cross-section creates both compressed stress and the bending moment, which causes the paper to bend upward.
\begin{figure} [t]
    \centering
    \includegraphics[width=\linewidth]{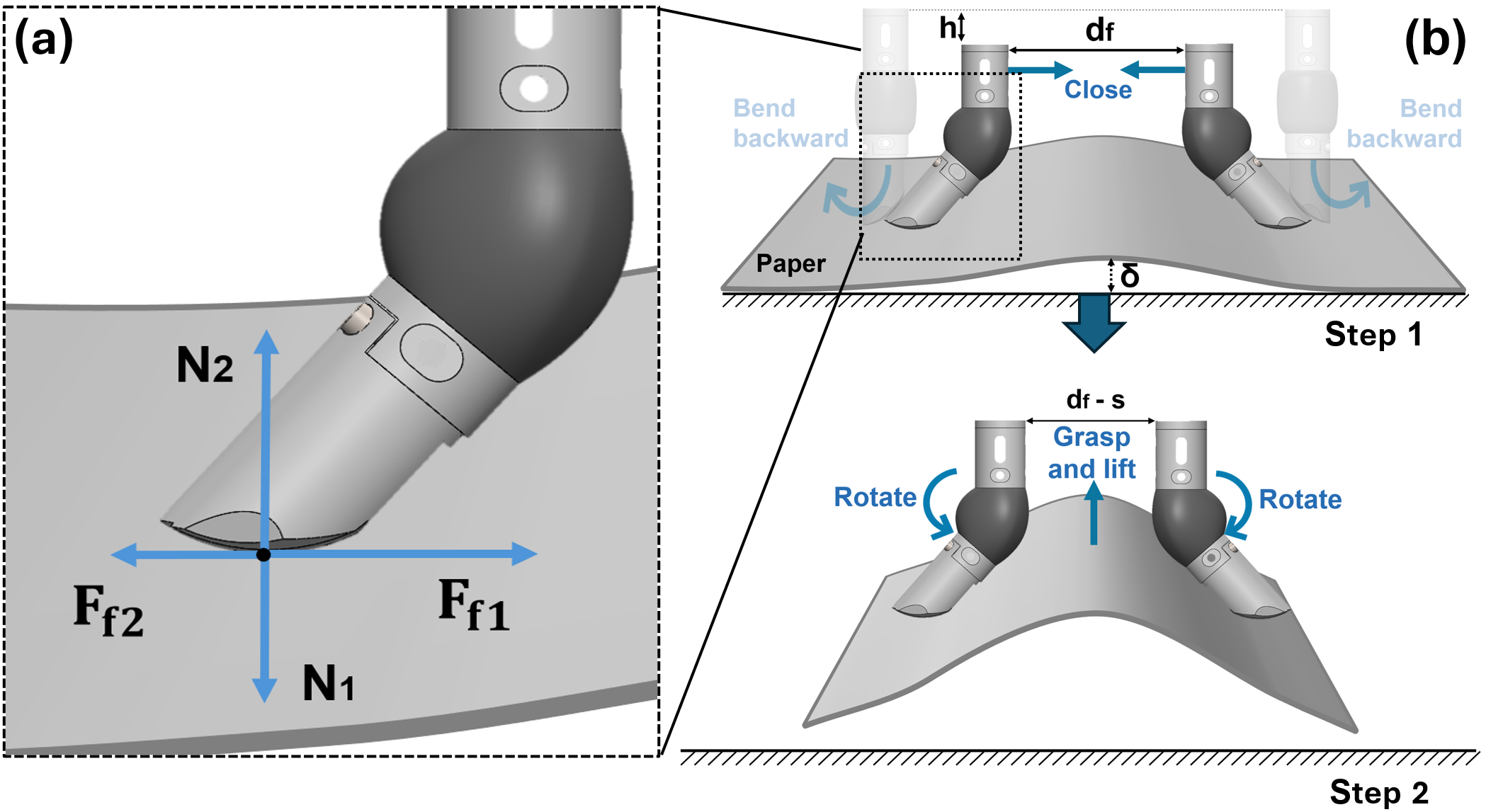}
    \caption{Grasping paper process. (a) Detail View expresses forces from fingertip and surface exerted on paper while the finger is closed. (b) Two steps of grasping paper using our gripper include closing two fingers with passive fingertip bend applying force on paper in step 1 and grasping up paper with the fingertip's rotating back which exerts gripping force on paper in step 2.  }
    \label{fig: grasping hypothesis}
\end{figure}

This experiment is designed to evaluate the significance of the gripper finger bending backward when grasping thin and wide objects, such as a sheet of paper. Theoretically, applying sufficient force at both ends of the paper sheet can cause it to bend, allowing it to be lifted. The greater the bend, the higher the friction between the gripper and the paper, enabling the paper to be grasped and lifted. If the paper is considered a beam, the bending stiffness of the paper $S_{b}$ can be calculated using the following equation \cite{mark2002handbook}, \cite{borch2001handbook}:
 \begin{equation}
S_{b}=\frac{M }{b(\frac{1}{R})}
\label{eq: bending stiffness 2}
\end{equation}
where $M$ is the bending moment applied to the paper, $\frac{1}{R}$ is the curvature of the paper, and $b$ is the width of the paper. Additionally, the bending stiffness of the paper can be approximated by the following equation \cite{mark2002handbook}, \cite{borch2001handbook}:
\begin{equation}
S_{b}=\frac{Et^3}{12}
\label{eq: bending stiffness 1}
\end{equation}
where $E$ is Young’s elastic modulus and $t$ is the thickness of the paper. According to this equation, the bending stiffness $S_{b}$ tends to be higher for the thicker paper which requires more bending moment to bend. 

When the gripper fingertip touches the paper, the force components can be modeled as shown in Fig.~\ref{fig: grasping hypothesis}(a). Here, $N_{1}$ is the force from the fingertip on the paper, and $F_{f1}$ is the frictional force between the fingertip and the paper; $N_{2}$ is the force from the surface exerted on the paper and $F_{f2}$ is the frictional force between the paper and the surface.

In this experiment, a paper sheet measuring $\SI{48}{\milli\meter}\times\SI{210}{\milli\meter}$ and \SI{0.5}{\milli\meter} in thickness is used. Similar to how a human would pick up a sheet of paper, the gripper fingers will press down on the paper's surface and gradually close in to lift the paper. The experiment is divided into two steps. In step~1, the two gripper fingers move downward to achieve a bending angle~$\alpha$ relative to the vertical axis of \SI{35}{\degree}, \SI{45}{\degree}, and \SI{65}{\degree}. This bending angle is achieved by adjusting the gripper’s displacement by a distance $h$, as follows:
\begin{equation}
h=d_1(\cos(\beta)-\cos(\beta+\alpha))
\end{equation}
The distance between the two gripper fingers in the first step is kept constant at \SI{30}{\milli\meter} for all trials. After that, a fixed amount of pressure is applied to the soft pneumatic ring in each case, with pressure values ranging from 0 to 150 kPa. In step~2, the gripper fingers close by \SI{5}{\milli\meter} to grasp and lift the paper sheet. For each $\alpha$ angle and pressure input, the grasping experiment is repeated 10 times, and the success rate of the paper grasping is recorded for evaluation. The result of this experiment is discussed in section~\ref{sec:result-grasp-paper}.

\subsubsection{Grasping \& manipulation of different objects} \label{sec: Grasp Objects}

This experiment involves grasping various objects with different materials, including a toilet paper roll, a plastic box, a PET bottle, a glue stick, and a charger (as shown in Fig.~\ref{fig: Introduction}), using the same experimental device as the grasping experiment with paper.

Regarding the experiment process, each test object was first placed on the 3D-printed basement. Then, the gripper's position was adjusted to attempt grasping with the same conditions in each trial, including finger distance and contact position, but with different pneumatic pressure. Each object was grasped 10 times for each pressure case, and the success of the grasping was recorded to calculate the success rate.  The result of this experiment is discussed in section~\ref{sec:result-grasp-objects}.
\subsubsection{Grasping paper experiment using hybrid and rigid gripper} \label{subsec: Experiment 3}

The rigid gripper finger in this experiment is similar to the proposed hybrid gripper finger but without the revolute joint between the distal and proximal links. The fingertip of the rigid gripper is identical to that of the proposed hybrid gripper. In this experiment, the hybrid gripper finger is initially positioned with a curvature angle of $\alpha$, the distance between the two contact points ($d$), and the Z-axis interaction force on the paper is $N_{1}$. The distance $d$, as shown in Fig. \ref{fig: finger and fingertip}(a), is calculated using the following equation:
\begin{equation}
    d=d_f+2d_e
\end{equation}
with $d_f$ standing for the distance between two fingers, $d_e$ standing for the extended distance from the contact point of the fingertip and finger body's axis:
\begin{equation}
    d_{e}=d_1\sin(\beta+\alpha)
\end{equation}
The rigid gripper will also undergo the same experiment, with the initial distance between the two contact points ($d$) and the z-axis interaction force on the paper ($N_{1}$). In both experiments, with the rigid and hybrid grippers, the gripper fingers will gradually close to grasp and lift the paper. The experiment is performed under three different conditions: 1st condition ($\alpha$ = \SI{35}{\degree}, $d$ = 60 mm, $N_{1}$ = 6 N), 2nd condition ($\alpha$ = \SI{45}{\degree}, $d$ = 65 mm, $N_{1}$ = 9 N), and 3rd condition ($\alpha$ = \SI{65}{\degree}, $d$ = 70 mm, $N_{1}$ = 14 N).The result of this experiment is discussed in section~\ref{sec:result-soft-rigid-comparison}.




\section{Experiment Results and Discussions}
\subsection{Fundamental experiment results}\label{sec:result-torque}
\begin{figure}
    \centering
    \includegraphics[width=.7\linewidth]{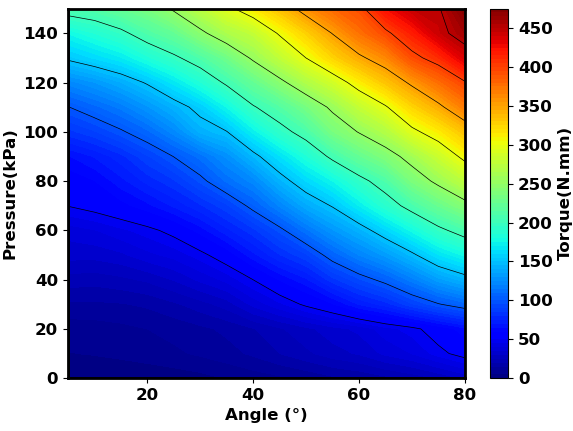}
    \caption{The torque generated by the soft pneumatic ring on the fingertip of the gripper.}
    \label{fig:result-joint-torque}
\end{figure}


\begin{figure}[t]
    \centering
    \includegraphics[width=1\linewidth]{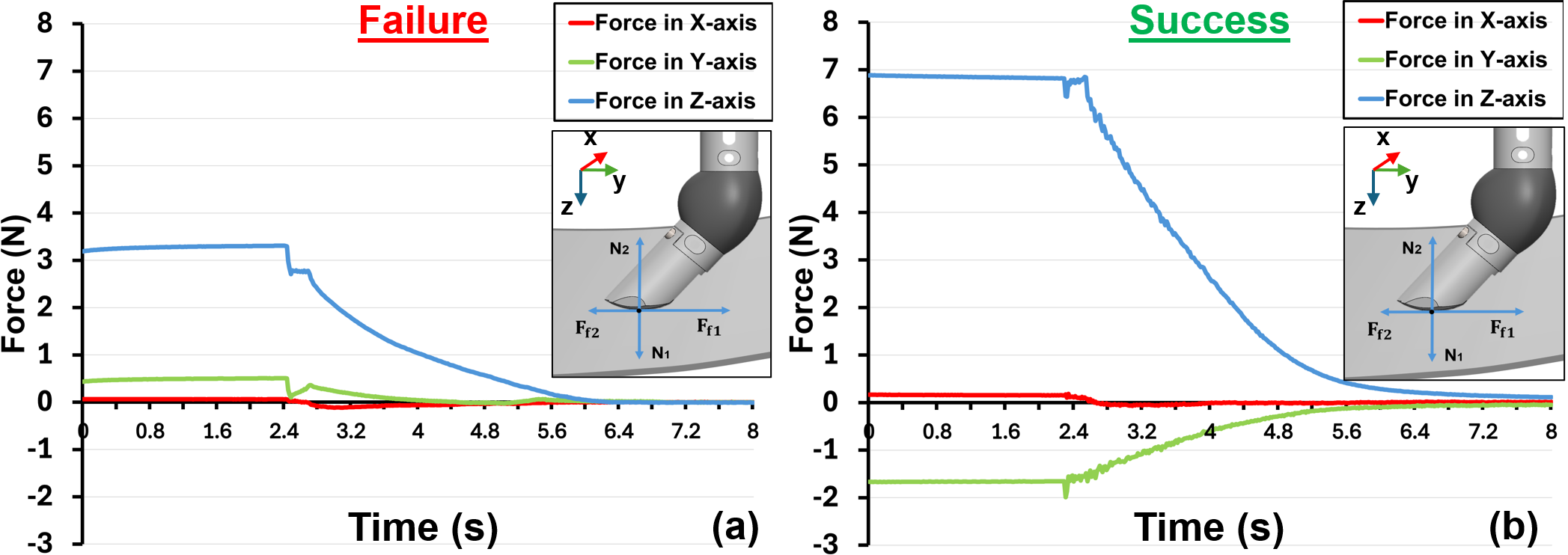}
    \caption{ Applied force during grasping paper sheet. (a) Force exerted from the fingertip on the surface in case of grasping failure ($\alpha =\SI{45}{\degree}$, zero-pressure input). (b) Force is exerted from the fingertip on the surface in case of grasping success, ($\alpha =\SI{45}{\degree}$, \SI{100}{\kilo\pascal} pressure input).}
    \label{fig:result-paper-grasp-force}
\end{figure}

Fig.~\ref{fig:result-joint-torque} shows that the torque generated by the soft pneumatic ring strongly correlated to the bending angle $\alpha$ and the input pressure. The increment of either the bending angle or input pressure will result in the increment of the resistance torque or the stiffness of the finger joint. 
Within the torque range of \qtyrange{150}{400}{\newton.\milli\meter}, the pressure and bend angle are relatively proportional. This indicates that with high pressure and a sufficiently large bending angle, the likelihood of successfully grasping objects will be higher. 

\begin{figure}[t]
    \centering
    \includegraphics[width=.9\linewidth]{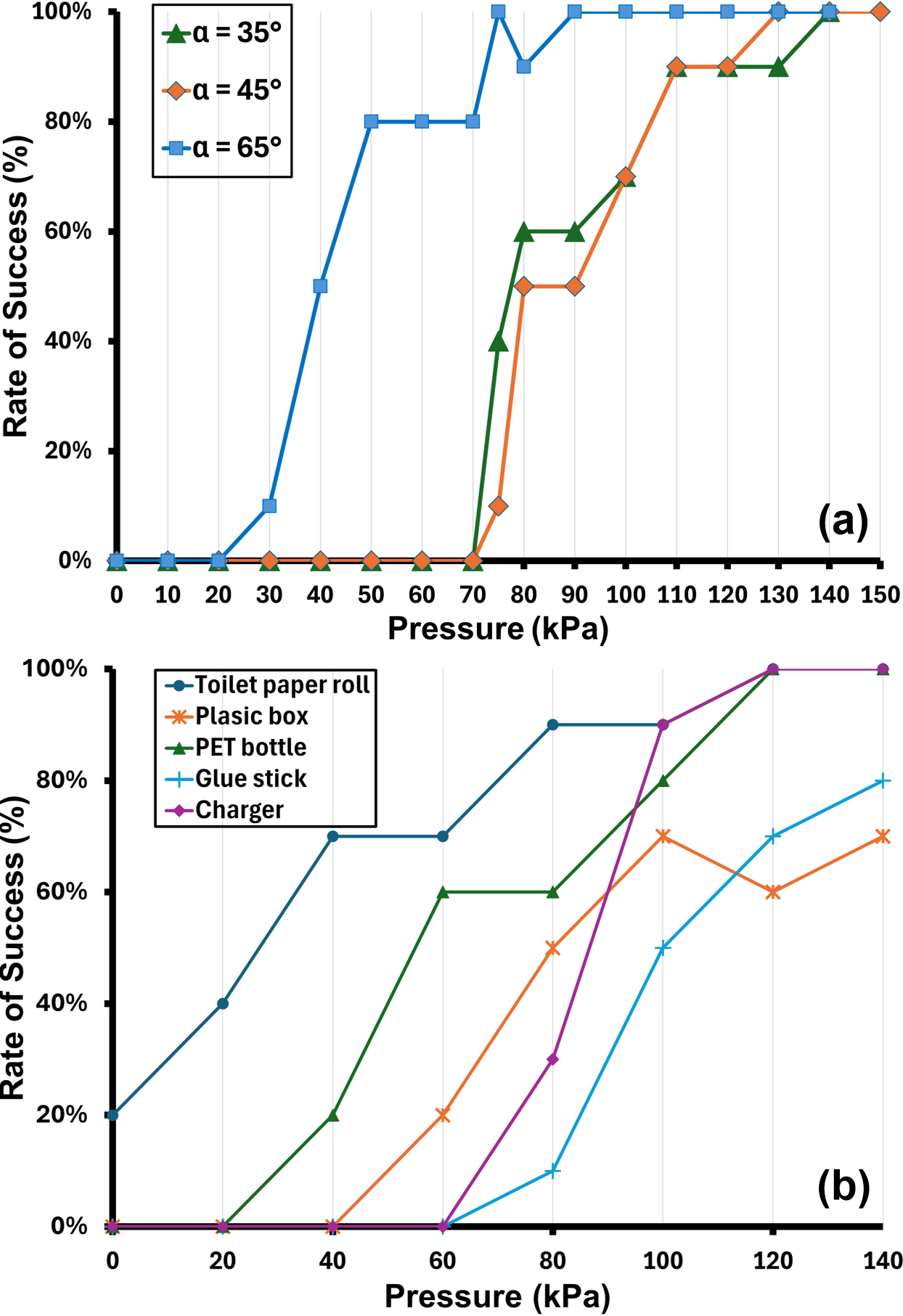}
    \caption{ Grasping objects successful rates. (a) Successful rate of soft gripper in grasping paper with different bending angles and pneumatic pressure. (b) The successful rate of the soft gripper in grasping five objects with different pneumatic pressure.}
    \label{fig: success rate different pressure}
\end{figure}

\subsection{Grasping experiment results}

\subsubsection{Thin-large object grasping experiment }\label{sec:result-grasp-paper}
 (Fig.~\ref{fig: success rate different pressure}(a)) shows the success rate in the paper grasping process with three different bending angles and varying air pressure levels supplied to the soft pneumatic ring. The results indicate that with a larger bending angle of $\SI{65}{\degree}$, even a small pressure input results in a high success rate, reaching $80 \%$ at a pressure of \SI{50}{\kilo\pascal}. Additionally, at a higher pressure of \SI{110}{\kilo\pascal}, the success rate of grasping the paper reaches $90 \%$, regardless of the bending angle. This can be explained by the fact that as the input pressure and angle increase, the resisting moment also increases, leading to a higher success rate in grasping objects.

\begin{figure*}[t]
    \centering
    \includegraphics[width=1\linewidth]{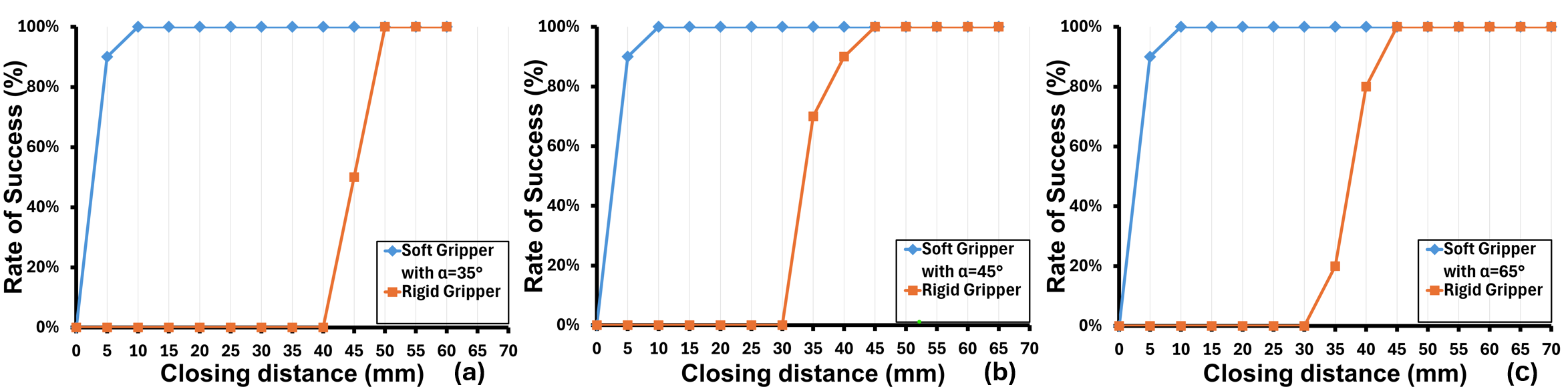}
    \caption{Successful rate of grasping objects of the rigid gripper with different closing distances measured under the same condition of the soft gripper (distance $d$ between two contact points and the normal force $N_1$ exerted on the surface) in case of (a) $\alpha=\SI{35}{\degree}$, $d =\SI{60}{\milli\meter}$, $ N_1=\SI{6}{\newton}$, (b) $\alpha=\SI{45}{\degree}$, $d =\SI{65}{\milli\meter}$, $N_1=\SI{9}{\newton}$, (c) $\alpha=\SI{65}{\degree}$, $d =\SI{70}{\milli\meter}$, $N_1=\SI{14}{\newton}$.}
    \label{fig: Success rate of rigid gripper compared to soft}
\end{figure*}
 The interaction force of the gripper on the paper, measured by the force sensor, is shown in Fig.~\ref{fig:result-paper-grasp-force}. The force values are observed from the moment when the normal force ($F_{z}$) from the gripper stabilizes and maintains at a certain value until the gripper lifts to attempt grasping the paper. The force values are recorded in two cases: when the object is successfully grasped (Fig.~\ref{fig:result-paper-grasp-force}(b), input pressure~=~\SI{100}{\kilo\pascal}) and when the object is not successfully grasped (Fig.~\ref{fig:result-paper-grasp-force}(a), input pressure~=~\SI{0}{\kilo\pascal}), both at the same initial bending angle of the fingertip of $\SI{45}{\degree}$. 

 The results show that in the case of a successful grasp, the $F_{z}$ force is greater than in the case of an unsuccessful grasp. Additionally, in the failure case, $F_y$ is small since the fingertip generates an insufficient friction force on the paper for bending, leading to the sliding between the fingertip and the paper. Ideally, in this case, paper and surface form a unit surface, so $F_y$ is only contributed by the friction force from the fingertip. In contrast, $F_y$ in successful grasping is higher but in the opposite direction due to the sufficient friction force exerted on the paper. Therefore, the friction force from the paper on the sensor is the main source for  $F_y$ when the paper is slid on the surface.

\subsubsection{Grasping different objects}\label{sec:result-grasp-objects}
In the cases of grasping a variety of objects (Fig.~\ref{fig: success rate different pressure}(b)), the proposed hybrid gripper gradually increases its grasping performance when the input pressure is increased. With a success rate of over 80$\%$ for the toilet paper roll, plastic box, and charger at an input pressure of \SI{100}{\kilo\pascal}, the results are promising. However, for objects like the plastic box or glue stick, the success rate only reached 70$\%$, even when the input pressure was increased to \SI{140}{\kilo\pascal}. As the pneumatic pressure for the soft ring increases, with the same gripping condition, the gripper applies more contact force, enabling objects to be lifted off the surface. However, with objects with sharp edges and small contact areas, the frictional force exerted by the fingertip is insufficient to trade off the gravitational force on the objects.

\subsubsection{Grasping paper experiment using hybrid and rigid gripper}\label{sec:result-soft-rigid-comparison}
Fig.~\ref{fig: Success rate of rigid gripper compared to soft} shows the performance of picking up a piece of paper between the proposed hybrid gripper and the rigid gripper under the same working conditions. The results indicate that the proposed hybrid gripper only needs to move \SI{5}{\milli\meter} to achieve a 90$\%$ success rate, while the rigid gripper requires a movement of about \SI{40}{\milli\meter} to achieve a success rate of 80-90$\%$. This demonstrates that, under the same working conditions, the proposed hybrid gripper performs better than the rigid gripper in grasping objects.

\subsection{Limitations \& Future works}

Although the experimental results show that the proposed hybrid gripper has a high potential for use in robotic applications requiring the grasping of various objects, especially those involving thin and large objects, the gripper still has some limitations. Currently, the study has only evaluated the role of bending the fingertip backward, and the effect of the fingertip bending inward has not yet been considered. Future research will explore the impact of inward fingertip bending on object grasping. Additionally, the soft pneumatic ring is currently separate from the gripper finger links to facilitate the study, but this makes the gripper bulky. In further research, the soft pneumatic ring will be integrated into the inside of the gripper finger to make the gripper more compact. Furthermore, other potential research could include developing Artificial Intelligent applications to adaptively control the air pressure supplied to the gripper finger to enhance grasping efficiency. Increasing the degrees of freedom of the gripper by adding more links is also another potential research.

\vspace{-0.1cm}
\section{CONCLUSIONS}
In this study, a hybrid robotic gripper is proposed, inspired by the way human hands grasp thin and wide objects when the distal phalanx tends to bend backward to grasp the object. The gripper consists of both rigid and soft components. The soft component can alter the stiffness of the gripper joint by increasing the input pressure. Fundamental results indicate that the bending angle, input pressure, and the resisting torque generated at the gripper joint are proportional to each other. Grasping experiments show that the proposed hybrid robotic gripper can successfully grasp objects, particularly thin and wide items like paper, with a high success rate under conditions of large bending angles and high input pressure. Another experiment also demonstrates that the proposed hybrid robotic gripper has a higher and more effective grasping success rate compared to a rigid gripper under the same conditions. These results suggest that the proposed hybrid robotic gripper has potential applications in various robotic applications. In future research, the hybrid robotic gripper will be designed to be more compact and intelligent to enhance its efficiency in grasping objects.

\addtolength{\textheight}{-9.5cm}   
\bibliographystyle{IEEEtran}
\bibliography{biblio}

\end{document}